\documentclass[conference]{IEEEtran}
\ifCLASSINFOpdf
\else
\fi

\usepackage{amsmath}
\usepackage{graphicx} 
\usepackage{array}    
\usepackage [autostyle, english = american]{csquotes}
\MakeOuterQuote{"}

\usepackage{url}

\begin{document}
%



\title{LLM Chain Ensembles for Scalable and Accurate Data Annotation}





%

\author{\IEEEauthorblockN{David Farr\IEEEauthorrefmark{1}\IEEEauthorrefmark{2},
Nico Manzonelli{2},
Iain Cruickshank\IEEEauthorrefmark{3},
Kate Starbird\IEEEauthorrefmark{1} and
Jevin West\IEEEauthorrefmark{1}}
\IEEEauthorblockA{\IEEEauthorrefmark{1}
University of Washington,
Seattle, WA}
\IEEEauthorblockA{\IEEEauthorrefmark{2} Army Cyber Technology and Innovation Center, Augusta, GA}
\IEEEauthorblockA{\IEEEauthorrefmark{3} Carnegie Mellon University, Pittsburgh, PA}
}


\maketitle

\begin{abstract}
The ability of large language models (LLMs) to perform zero-shot classification makes them viable solutions for data annotation in rapidly evolving domains where quality labeled data is often scarce and costly to obtain. However, the large-scale deployment of LLMs can be prohibitively expensive. This paper introduces an LLM chain ensemble methodology that aligns multiple LLMs in a sequence, routing data subsets to subsequent models based on classification uncertainty. This approach leverages the strengths of individual LLMs within a broader system, allowing each model to handle data points where it exhibits the highest confidence, while forwarding more complex cases to potentially more robust models. Our results show that the chain ensemble method often exceeds the performance of the best individual model in the chain and achieves substantial cost savings, making LLM chain ensembles a practical and efficient solution for large-scale data annotation challenges.
\end{abstract}


%
\IEEEpeerreviewmaketitle

\section{Introduction}

Quality-labeled data is fundamental to the development of machine learning (ML) models, which facilitate advanced data analysis in decision-making processes across both research and industry. Obtaining structured data with accurate labels presents significant challenges, particularly in dynamic fields like Computational Social Science (CSS), where classification objectives are less defined and may change frequently due to current events. Additionally, traditional data annotation methods that rely on human annotators are expensive and time-consuming making them impractical for rapidly evolving domains that require massive datasets. In response, there is a growing trend of using large language models (LLMs) for data annotation \cite{ziems24, redct, zhu2023can}.

LLMs are a promising solution to fields with dynamic data annotation requirements due to their ability to classify instances that were previously unseen in training data, also known as zero-shot classification \cite{wang2019survey}. Zero-shot classification offers advantages in speed, scalability, and flexibility. However, despite these benefits, the use of LLMs also introduces new complexities. As the volume of data increases, so do the computational demands and associated costs of deploying high-parameter LLMs. Furthermore, the financial burden of accessing these models through commercial APIs can become prohibitively high, particularly for large-scale annotation tasks. Finally, zero-shot annotation by LLM alone is usually insufficient to produce a high-quality, labeled dataset \cite{cruickshank2024promptingfinetuningopensourcedlarge, zhang2023llmaaa}.

In this paper, we introduce an LLM chain ensemble method to address the challenge of obtaining accurate labels for text data in a time, computational resource, and cost-efficient manner. Previous work in text classification for CSS tasks using LLMs focuses mainly on tasks such as sentiment, stance, ideology, or misinformation detection \cite{
ziems24, cruickshank2023use, cruickshank2024promptingfinetuningopensourcedlarge}. These works place an emphasis on prompt engineering techniques and evaluating performance comparisons across different LLMs. Our approach marks a significant departure from this trend. Rather than focusing on prompt engineering or comparing LLMs, we treat individual LLMs as integral components within a broader system designed to enhance classification performance and reduce costs.

Our proposed methodology aligns LLMs in a sequence, routing subsets of the data that each LLM is uncertain about to subsequent models in the sequence based on zero-shot classification uncertainty measures. This approach enables each LLM in the chain to label the data in which it is most confident, while forwarding more challenging instances to subsequent, potentially more robust models. After each LLM-link assigns labels to its data, the predictions and their corresponding confidence scores are aggregated using a rank-based ensemble method to derive the final labels. This hierarchical structure allows earlier, ideally, less costly or resource-intensive LLMs, to handle easier examples to classify, reserving more complex cases for later stages in the chain that may benefit from an ensemble.

\begin{figure}
    \centering
    \includegraphics[scale=.48]{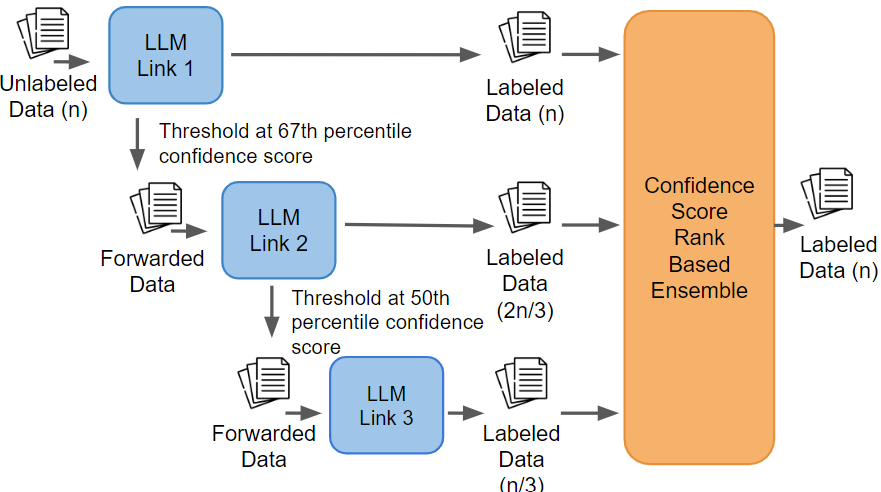}
    \caption{This is a system diagram of our chaining methodology. The system depicts routing paths for subsets of data to pass to subsequent LLMs informed by a calculated confidence metric. After all data has been labeled by at least one LLM, we assign final labels informed by our rank-based ensemble.}
    \label{chain}
\end{figure}

Our chain ensemble approach demonstrates the capacity to exceed the performance of the most accurate or expensive individual model in the chain, despite using that model to label only a fraction of the dataset. Our approach improves on existing benchmarks across three distinct zero-shot CSS tasks using straightforward prompting techniques.

In contrast to more intricate prompting methods, such as chain-of-thought or tree-of-thought, our approach deliberately minimizes token usage by constraining LLM responses to a single word. This strategy not only increases classification speed but also offers substantial cost savings, delivering up to a 90-fold reduction in expenses compared to other complex prompt engineering techniques employed with commercial LLMs, all while maintaining comparable or superior performance. This combination of cost-efficiency, speed, and high performance underscores the practicality and effectiveness of our chain ensemble methodology in large-scale data annotation tasks.


\section{Related Works}

Prior research shows that supervised ML solutions often fail to generalize to out-of-domain data, making them difficult to implement in dynamic fields like CSS where quality-labeled data is scarce \cite{NG2022103070}. Due to the impressive ability of LLMs to perform zero-shot predictions, they have garnered significant attention as viable alternatives for text classification or data annotation in CSS \cite{ziems24, cruickshank2023use, zhu2023can, he2023annollm, wang-etal-2021-want-reduce}. While LLMs may not consistently outperform in-domain supervised models, they remain competitive, particularly in scenarios where domain-specific data is scarce or rapidly changing \cite{cruickshank2024promptingfinetuningopensourcedlarge}. Recent research suggests that LLMs can serve as an effective solution for data labeling, especially in cases where annotated data is unavailable, the data-generation process evolves too quickly for traditional model development methodologies to keep pace, or the cost of training in-domain supervised classifiers is prohibitive \cite{redct}. Much work is available in literature on using LLMs for the purpose of data annotation \cite{tan2024largelanguagemodelsdata}; however, our work takes a novel approach with our chaining mechanism based on our designed confidence score metric to determine data routing to subsequent chain links.

Studies consistently demonstrate that no single LLM dominates across all natural language processing (NLP) tasks; instead, different LLMs exhibit varying levels of performance depending on the specific task at hand \cite{ziems24, gatto2023chainofthoughtembeddingsstancedetection, xu2024llmsclassificationperformanceoverclaimed}. This variability in performance has led to the development of specialized prompt engineering techniques, such as COLA or Chain of Stance, which have been shown to enhance classification outcomes \cite{wei2023chainofthoughtpromptingelicitsreasoning,gatto2023chainofthoughtembeddingsstancedetection, ma2024chainstancestancedetection, lan2024stancedetectioncollaborativeroleinfused}. Our work is independent of specific prompting techniques and diverges from this line of research by focusing on a multi-LLM classification system that aggregates confidence scores across models.

LLM ensemble methods combine LLMs in an effort to improve reliability and performance on downstream tasks. In contrast to popular Mixture of Experts (MoE) methods, which focus on learning MoE layers and routing layers from scratch, LLM ensemble systems combine publicly available pre-trained LLMs \cite{adaptivemixturesoflocal,eigen2014learningfactoredrepresentationsdeep, jiang2024mixtralexperts}. Despite the improvement in performance offered by using many LLMs for annotation, ensemble methods that aggregate LLM outputs after inference are often infeasible for large-scale data annotation due to their high costs: each model in the ensemble must perform inference on every example \cite{blender}. 

To reduce the labeling costs of using ensembles of models, researchers have considered learning external routing models for LLM ensembles \cite{shnitzer2023largelanguagemodelrouting, lu2023routingexpertefficientrewardguided, ding2024hybridllmcostefficientqualityaware, chen2023frugalgptuselargelanguage}. First explored by \cite{shnitzer2023largelanguagemodelrouting}, LLM routing methods attempt to learn a model that can map queries to candidate LLMs. \cite{lu2023routingexpertefficientrewardguided} shows promising results for LLM routing, but their method is limited by the potential cost of routing all queries to a few expensive LLMs. \cite{ding2024hybridllmcostefficientqualityaware} aim to decrease inference costs by learning a routing model that predicts the difficulty of a query and maps easier prompts to smaller, less expensive models and harder queries to more complex models. While these methods have the potential to reduce costs for large-scale data annotation, they may underperform compared to other LLM ensembling methods because they do not compare or aggregate outputs from multiple LLMs \cite{srivatsa2024harnessingpowermultipleminds}. Additionally, these methods require training routing models, which may be costly or time-consuming in its own right.

Finally, weak supervision has also been incorporated with ensembles of models to both improve label quality and potentially reduce labeling costs. \cite{smith2024language} showed how to combine weak supervision with prompting to improve LLM-derived labels. In \cite{huang2024alchemist}, the authors demonstrated that an LLM could be used to generate weak labeling functions --- which are incorporated into weak supervision --- instead of labeling the points directly. While these methods do improve ensembles and show some potential at reducing cost, it is unclear whether these methods work for CSS tasks and they do not recognize the per-instance hardness differences present in many tasks that could allow for resource savings.

Our proposed LLM ensemble method takes a similar design approach to \textit{FrugalGPT} \cite{chen2023frugalgptuselargelanguage}, which aligns LLMs in a chain and routes subsets of data that do not meet a predefined ``reliability score" threshold to the next LLM in the chain.\footnote{The authors refer to this sequential design as a LLM cascade instead of a chain.} However, our solution differs significantly. First, their ``reliability score" requires subsets of in-domain labeled data, whereas our LLM chain ensemble uses a generalized confidence score for zero-shot predictions, eliminating the need for any prior data labeling. Furthermore, while their method discards the labels and scores of prior LLM predictions, our approach retains all available scores in a rank-based ensemble to select the best label. Finally, our work uniquely focuses on scalability and data annotation for complex data domains and tasks presented by CSS, where traditional routing or ensemble methods may not be as effective.

Overall, the combination of scalability, cost-efficiency, and adaptability makes our LLM chain ensemble a robust alternative to existing methods, particularly in scenarios where rapid and accurate data annotation is critical.

\section{LLM Chain Ensembles}

In this section, we introduce the LLM chain ensemble, which treats pre-trained LLMs as components of a classification system. By leveraging multiple LLMs in a sequential manner, our approach aims to enhance classification accuracy while decreasing computational costs.

We refer to the sequential alignment of LLMs as a “chain” and each LLM pipeline within the chain as a “link”. At each link \( f_{L_i}\), raw text data is converted into a prompt and passed through the LLM to obtain a label and a corresponding confidence score. We seek to label a dataset \( X = \{x_1, \dots, x_n\} \) with an LLM chain \(\mathbf{L} = \{f_{L_1}, \dots, f_{L_m}\} \) by sequentially passing subsets of the data through each link such that later LLM links see smaller proportions of the dataset. 

The LLM chaining process starts by labeling all observations with the first link \( f_{L_1} \). Then, for each subsequent link $i \in \{1, \dots, m\}$, we select the top \(\frac{m-i+1}{m}\) fraction of labeled observations based on their confidence scores to retain at the current link, while forwarding the remaining observations to the next link in the chain. These specific fractions were chosen to ensure that higher-confidence predictions are retained early on, reducing the load and potential redundancy in subsequent LLMs. For example, with a chain length of \( m = 3 \), this process is as follows: link 1 labels 100\% (\( n \)) of the observations, link 2 labels \(\frac{2}{3}\) (\( \frac{2n}{3} \)) of the observations and link 3 labels \(\frac{1}{3}\) (\( \frac{n}{3} \)) of the observations, as shown in Figure \ref{chain}.

\subsection{Estimating Model Confidence}

To decide which portion of the data gets labeled in subsequent chain links, we quantify the confidence in the LLM's predictions. We compute a confidence score based on the token label log probabilities returned by each LLM. This method is inspired by approaches demonstrated in \cite{redct}, which showed the efficacy of using log probabilities to gauge prediction confidence. However, we apply a slight variation by constraining the set of token label log probabilities to only those corresponding to the set of expected labels, \(\mathcal{T}\), and then build on it with a ranking and normalization mechanism. 

Specifically, the confidence score is defined as the absolute difference between the highest token label log probability and the second-highest token label log probability within this constrained set of expected tokens. Let \(\mathcal{T}\) represent the set of given tokens, and \(P(t)\) denote the distribution of log probabilities across each token \(t \in \mathcal{T}\). The confidence score, denoted as \(C\), is then computed using the formula

\begin{equation}
    \text{\(C\)} = \left| \max_{t \in \mathcal{T}} P(t) - \max_{t \in \mathcal{T} \setminus \{t^*\}} P(t) \right|,
\end{equation}

\noindent where \(t^*\) is the token corresponding to the highest probability \( \max_{t \in \mathcal{T}} P(t) \). In contrast 

\subsection{Forwarding Data}

At each chain link we select the top \(\frac{m-i+1}{m}\) fraction of labeled observations based on their confidence scores to retain at the current link, while forwarding the remaining observations to the next link. By only forwarding a fraction of the data at each link, we limit costs associated with using LLMs later in the chain. Additionally, we allow for improved prediction for routing examples the current LLM link is not confident about to the next LLM link in the chain. 

\begin{figure*}[]
    \centering
    \includegraphics[scale=.50]{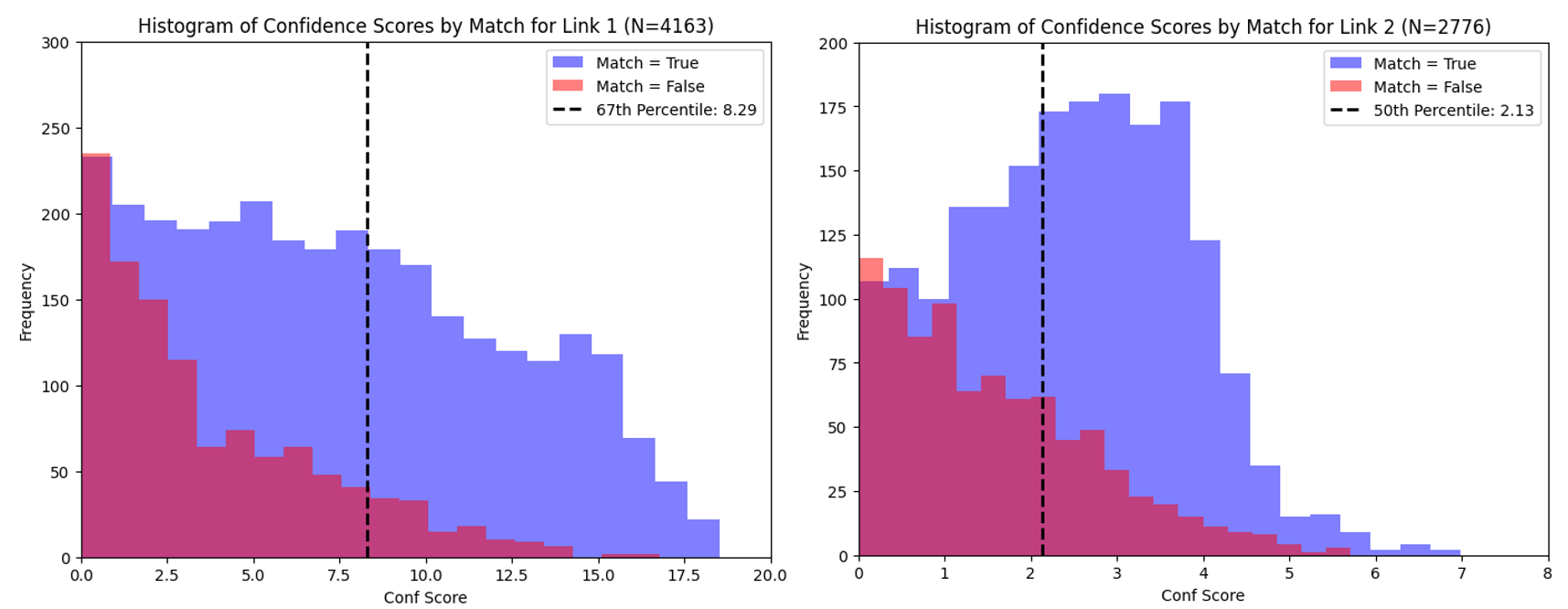}
    \caption{Confidence Score distribution at chain link 1 and chain link 2 for stance detection task stratified by correctly and incorrectly labelled examples. All data greater than the given threshold (67th percentile dashed line at link 1 and 50th percentile dashed line at link 2) is data retained at the current chain link for classification in the rank-based ensemble. Data to the left is forwarded to the next chain for future additional classification. The distributions shown are overlaid histograms to show the calculated confidence score when stratified by assigned labels that are true in blue and assigned labels that are false in red. }
    \label{fig:stance_forwarding}
\end{figure*}

Our data forwarding strategy is best at limiting costs when the chain sequence is ordered from cheapest to most expensive LLMs. Our overall assumption is that simple, cheap models can classify simple examples correctly, while leaving the more difficult annotation tasks to more complex models further down the chain. Figure \ref{fig:stance_forwarding} validates this assumption for a small sample by depicting the confidence score distributions across incorrectly and correctly labeled data on the stance detection task. All data to the left of the dashed line, representing our confidence score threshold, is forwarded to the next link. Figure \ref{fig:stance_forwarding} shows the utility in our data forwarding strategy, where the density of correctly labeled data is higher for data retained at the current link and lower for data forwarded to the next link.

While using thresholds based on confidence score percentiles is effective, it is important to recognize that the threshold for forwarding data can be adjusted depending on the specific use case. In scenarios where factors such as cost, computational resources, or time are critical, it might be better to further tune the amount of data that is forwarded. Additionally, our thresholding method operates under a data batch processing framework. After an initial sample of data has been processed, it is possible to dynamically adjust the forward thresholds based on new data or learn these thresholds once and apply them to new data.

\subsection{Rank Based Ensemble}

To capitalize on the cumulative knowledge from all previous LLM data labeling efforts in the chain, we retain each links' classification and confidence score for a rank based ensemble. At each link $f_{L_i}$, we rank and normalize the confidence scores $C_j$ for all $j$ observations classified by the link. Then, we sort the confidence scores such that $\mathsf{rank}(C_j) = 1$ for the lowest score and $\mathsf{rank}(C_j) = n_i$ for the highest score where $n_i$ is total number of data points classified at link $i \in \{1, \dots, m\}$. link. The normalized rank for each data point is computed as

\begin{equation}
R_j = \frac{\mathsf{rank}(C_j)}{n_i},
\end{equation}

\noindent which creates a uniform distribution over $R_j$ constrained to \([1/n_i, 1]\).

Ranking and normalization allows confidence scores from different LLMs to be compared directly, regardless of differences in LLM prediction which may impact confidence scores. Instead of simply adopting the labels retained at each link, our approach selects the label with the highest normalized ranking value across all available prior links. Formally, for all $n_i$ observations labeled at link $f_{L_i}$, let $y_{j,i}$ be the label assigned by link $f_{L_i}$ and $R_{j,i}$ be the corresponding normalized ranking value where $i \in \{1, \dots, m\}$ and $j \in \{1, \dots, n_i\}$. The final label for observation $y^*_j$ is defined as

\begin{equation}
    \label{eq:argmax}
    y^*_j = y_{i^*}, \quad \text{where } i^* = \arg \max_{i \in \{1, \ldots, m\}} R_{j,i}.
\end{equation}

Under our rank based ensemble, only the classification from the first LLM is considered at link $i = 1$ so $y^*_j = y_{j,1}$. At link 2, the classifications of links 1 and 2 ($y_{j,1}$ and $y_{j,2}$) are evaluated with Eq. \ref{eq:argmax}, and this process continues for each subsequent link in the chain. By considering all available labels and their relative confidences, we leverage the strengths of specific LLMs that may excel at particular tasks or on certain observations. The rank ensemble ensures that the final label selected is the label associated with the highest relative confidence across all available links.

Running the LLM chain with the rank based ensemble is particularly advantageous when a benchmark dataset is unavailable, and it is unclear which LLM is best suited for a given task. By integrating the strengths of multiple LLMs, our classification system adapts dynamically, ensuring the best possible label at each stage. This adaptability is crucial in real-world applications, where the diversity of tasks and data points often means that no single LLM can consistently outperform others. Considering the collective confidence of all models in the chain, our methodology provides a robust and flexible solution for data labeling in complex environments.

\section{Experimental Design}

Our research focuses on three text classification tasks that have readily available datasets and zero-shot classification benchmarks: stance detection, ideology detection, and misinformation detection. We evaluate each task with five LLMs in various chain permutations. Though these tasks are broadly similar in their goal of extracting meaningful information from data, each presents unique challenges and nuances.

\subsection{Stance Detection}

Stance detection is a classification problem where the objective is to determine the author's position on a particular topic, categorized as {Favor, Against, Neither} \cite{StanceDetDEf}. Detecting stance has significant applications in industry, particularly in understanding consumer opinions and conducting market research. By accurately identifying the stance of individuals or groups, businesses can better tailor their marketing strategies, improve customer satisfaction, and anticipate market trends. For instance, analyzing public sentiment towards a new product or a controversial policy can guide decision-making processes. We use the SemEval-16 dataset which is common across stance detection studies and provides us with well-established benchmarks \cite{semevaldata}.

\subsection{Misinformation Detection}

Misinformation detection involves identifying "false or inaccurate information that is deliberately created and intentionally or unintentionally propagated" \cite{misinfodefined}. In a corporate context, especially during a crisis, the ability to swiftly detect and mitigate misinformation is crucial for maintaining public trust and protecting brand reputation. Misinformation can lead to widespread public confusion, unfounded fear, or misguided actions that may have direct implications for a company’s products, services, or public image.

To validate our system’s effectiveness in detecting misinformation, we use the Misinfo Reaction Frames corpus \cite{misinfodata}. The dataset contains 25k news headlines on topics such as COVID-19, climate change, and cancer. Each headline was fact-checked and labeled with a binary classification of misinformation or trustworthy, providing a robust foundation for testing and refining misinformation detection tools that can be deployed in real-world industry scenarios.

\subsection{Ideology}

We define ideology as "the shared framework of mental models that groups of individuals possess, providing both an interpretation of the environment and a prescription as to how that environment should be structured" \cite{ideologydefined}. In various contexts, including public health and other critical areas, understanding consumer ideology can be instrumental in tailoring communication to make quality information more accessible and persuasive. When messaging aligns with an individual's ideological stance, it can enhance receptiveness and trust in the information source. 

We use the Ideology Books Corpus (IBC) dataset from \cite{sim-etal-2013-measuring}, complemented with sub-sentential annotations from \cite{iyyer2014neural}. The IBC contains 1,701 sentences with a conservative political leaning, 600 sentences labeled as neutral, and 2,025 sentences with a liberal political leaning. This dataset provides a robust framework for analyzing how ideological leanings can be detected and leveraged to improve communication strategies, particularly in disseminating important information.

\subsection{Candidate LLMs and Configurations}

We tested the LLM chain ensemble methodology using four open-source LLMs, LLAMA 3.1 (8B), Flan-UL2, Mistral-7b, and Phi3, as well as one closed-source LLM, GPT-4o.\footnote{Model sources are contained in the Ethics and Availability section.} These models offer a broad perspective on performance across various conditions because they provide a diverse range of parameter sizes, architectures, and cost profiles. To ease computational overhead and because implementing longer chains may be unrealistic in practice, we tested all permutations of candidate LLMs with chains of length 4 or less. Additionally, we propose two realistic \textit{production chains} that allow simpler models to label easier observations, while more complex and resource-intensive models address data points that achieve low confidence for earlier links in the chain.

\subsection{Prompting}

While our paper focuses on an ensembling method rather than prompting schemes, it is important to address the constraints related to prompting. Our methodology is agnostic to the specific prompt technique but not to the prompts themselves. For the methodology to be effective, the prompt must restrict the LLM's final output to a label within a predefined set of expected labels. Techniques such as chain-of-thought, tree-of-thought, and chain-of-prompt can be employed, provided they constrain the final LLM response to classification targets. For our experimentation, we chose the simple prompting techniques and used the same prompt across LLMs to avoid potential prompt over-fitting issues.

\subsection{Experimental Set-Up}

To eliminate the need to run all chain permutations for each dataset from scratch (1800 experiments from 600 permutations), we labeled each dataset once with each candidate LLM and simulate the chain ensemble method. Consequently, we only needed to retrieve labels for each dataset and each model once. With access to LLM labels for each model and task, we can easily compare the performances of the chain ensemble to random data forwarding strategies and individual models. 

We evaluate classification performance using the F1-macro score. To evaluate the impact of chain length and model order, we report average performances for all permutations of chains up to length four. We also report the averages from the simple forward chain method which keeps the labels from the last LLM link without the rank ensemble and random forwarding which randomly selects subsets of the data to forward to the next link. Finally, we report performance of each production chain compared to the best and worst performing LLM for each task.

\begin{figure*}[]
    \centering
    \includegraphics[scale=.61]{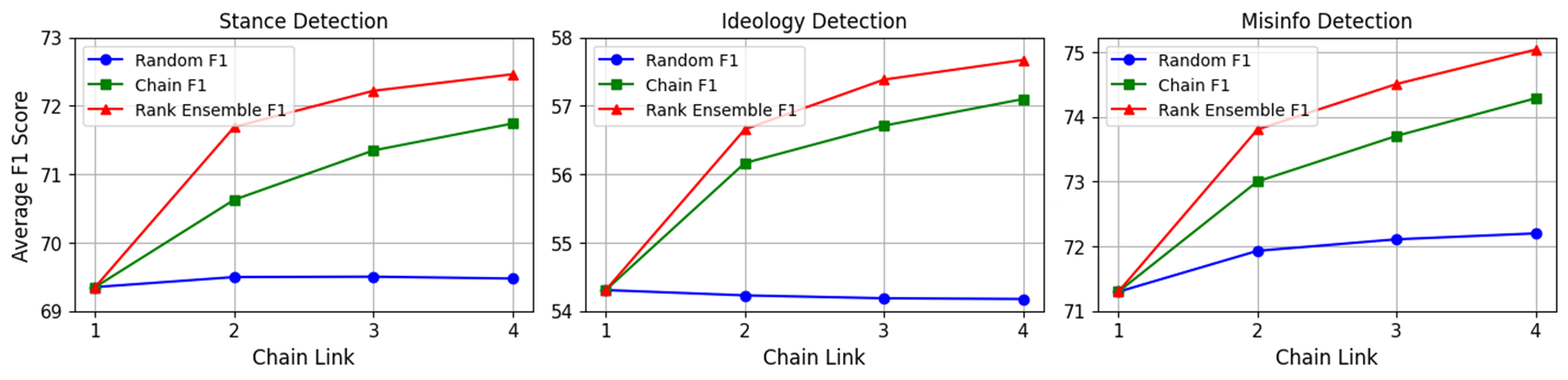}
    \caption{Depicts the average F1 performance across chaining technique at chain links 1 through 4 for stance, ideology, and misinformation tasks. The figure denotes a clear increase in F1 performance over random data forwarding across LLMs when using our confidence forwarding metric and additional performance increase when incorporating rank-based ensembling.}
    \label{chain_size}
\end{figure*}

\begin{table}[ht]
\centering
\label{Results}
\caption{Average F1 and Standard Deviation across Tasks for Individual LLMs, Forward Chain only, and Chain Ensembles of Length Four.}
\begin{tabular}{lcccccc}
\hline
 & \multicolumn{2}{c}{Stance} & \multicolumn{2}{c}{Ideology} & \multicolumn{2}{c}{Misinfo} \\
 & F1 & $\sigma$ & F1 & $\sigma$ & F1 & $\sigma$ \\
\hline
AVG LLM            & 69.35  & 7.89  & 54.31  & 8.23  & 71.30  & 10.32 \\
AVG Forward Chain  & 71.74  & 3.62  & 57.10  & 3.71  & 74.29  & 5.58  \\
AVG Chain Ensemble & 72.46 & 3.90  & 57.67  & 3.43  & 75.04  & 3.88 \\
\hline
\end{tabular}
\end{table}

\section{Results}

Figure \ref{chain_size} shows the averages of base model performance, random data forwarding (Random F1), forward chain only (Chain F1), and full chain ensemble (Rank Ensemble F1) across chains of sizes 2, 3, and 4, demonstrating improvement across chain length and chaining technique for each task. Chain size 1 is the average of all LLMs evaluated.

Table \ref{Results} reports the average F1 score and standard deviation across the 120 permutations of chains of length four. Compared to single LLMs or forward chain only, we see that the chain ensemble increases performance and decreases standard deviation across average F1s when compared to individual LLMs and forward chaining only.

In the following sections we breakdown our results by task and focus on two realistic, representative chain configurations: \textit{Production Chain 1} and \textit{Production Chain 2}. The LLMs in these chain are ordered by parameter size which acts a simple heuristic for cost and expected performance. Production Chain 1 is LLAMA 3.1, Flan-UL2, and GPT-4o. Production Chain 2 is Mistral-7b, Phi, Flan-UL2, and GPT-4o. We chose the LLMs for Production Chain 1 because of their consistent performance in all tasks, and the LLMs in Production Chain 2 because it includes models that performed poorly, namely Mistral-7b and Phi.

It's important to highlight the cost savings of our chaining implementation compared to chain-of-though prompting. We found that GPT-4o uses approximately 295 output tokens per response with a required input of 35 tokens on average for chain-of-thought prompting. When using Production Chain 1 and simple prompting, we only submit 1/3rd of the data to GPT-4o with an average input length of 25 tokens and max output length of 2 tokens. The price of using chain-of-thought prompting with GPT-4o to label 10 million observations would be approximately \$46,000 under current pricing. The costs associated with GPT4-o under Production Chain 1 would be approximately \$516 (90x cheaper not including the costs associated with running local models). Furthermore, while we delve into each classification task to better explore the nuances of performance, we do see results across all evaluated datasets that demonstrate improvement in data annotation performance and reduction in standard deviation of models by implementing our chain-ensemble methodology.

\subsection{Stance Detection}

\begin{table*}[ht]
\centering
\setlength{\tabcolsep}{4pt}
\label{stance-results}
\caption{LLM Chain Ensemble performance by chain link for Production Chain 1. Descending $F1$ and $F_{avg}$ scores across chain link show stratification of easy to harder samples to label despite models becoming increasingly complex. Chain link 1 is LLAMA, chain link 2 is Flan, and chain link 3 is GPT.}
\begin{tabular}{c|ccc|ccc|ccc|ccc|ccc|ccc}
\multicolumn{1}{c|}{} & \multicolumn{3}{c|}{LA} & \multicolumn{3}{c|}{CC} & \multicolumn{3}{c|}{AT} & \multicolumn{3}{c|}{HC} & \multicolumn{3}{c|}{FM} & \multicolumn{3}{c}{Total} \\ \hline
\textit{} & \textit{$F1$} & \textit{$F_{avg}$} & \textit{n} & \textit{$F1$} & \textit{$F_{avg}$} & \textit{n} & \textit{$F1$} & \textit{$F_{avg}$} & \textit{n} & \textit{$F1$} & \textit{$F_{avg}$} & \textit{n} & \textit{$F1$} & \textit{$F_{avg}$} & \textit{n} & \textit{$F1$} & \textit{$F_{avg}$} & \textit{n} \\ \hline
Chain Link 1& 76.9& 73.8& 281& 81.1& 87.1& 158& 72.0& 66.7& 114& 89.7& 93.2& 436& 68.9& 86.2& 398& 86.6& 91.0& 1387\\ 
Chain Link 2& 79.8& 79.1& 348& 72.8& 64.3& 189& 81.9& 85.6& 372& 75.3& 74.4& 276& 70.4& 68.5& 203& 82.1& 82.7& 1388\\ 
Chain Link 3& 67.8& 71.1& 304& 54.9& 49.3& 217& 58.6& 59.7& 247& 65.1& 65.0& 272& 65.2& 65.1& 348& 65.3& 66.2& 1388\\ \hline
\textbf{Full Chain}& \textbf{77.7}& \textbf{78.9}& 933& 71.2& \textbf{68.2}& 564& 77.3& \textbf{80.9}& 733& 79.1& \textbf{82.1}& 984& 73.6& \textbf{78.1}& 949& \textbf{78.2}& \textbf{81.02}& 4163\\
\end{tabular}
\end{table*}

For stance detection we include the recommended benchmark metric of SemEval-16 Task A $F_{avg}$, the macro F1 across the five target classes included in the SemEval-16 dataset (CC-Climate Change, AT-Atheism, HC-Hillary Clinton, LA-Legalization of Abortion and FM-Feminist Movement), and well as the overall performance of Production Chain 1. We show the performance across all target classes and two reporting metrics ease of comparison with other studies \cite{ziems24, aiyappa2024benchmarkingzeroshotstancedetection, ma2024chainstancestancedetection}. We further break our results down by chain link stage to show the effective stratification of easier to label examples based on confidence scores in earlier chain links. Our methodology breaks recently published benchmarks in \cite{ziems24, aiyappa2024benchmarkingzeroshotstancedetection, ma2024chainstancestancedetection}, which used various prompting techniques on GPT-4, Mistral, Llama 3, and Flan-T5 XXL. These expanded results are shown in Table \ref{stance-results}.

Table \ref{Results} shows the both the slight increase in average F1 comparing chain permutations against individual LLMs as well as the significant increase in model robustness shown by the greatly reduced standard deviation value using our chain ensemble method. Table \ref{stance-performance-comparison} highlights the performance of Production Chain 2 which outperforms still the worst performing LLM in the chain, highlighting the ability for chain ensembling to mitigate the impact of low-performing LLMs in a data annotation system while still saving significant resources.

\begin{table}[ht]
\setlength{\tabcolsep}{4pt}
\centering
\caption{Stance Detection Results}
\label{stance-performance-comparison}
\begin{tabular}{lcc}
\hline
\textbf{Category} & \textbf{Rand. Forward F1} & \textbf{Chain Ensemble F1} \\
\hline
Best Single LLM (GPT)       & -      & 76.75 \\
Worst Single LLM (Mistral)  & -      & 56.41 \\
Prod Chain Ensemble 1       & 73.93  & 78.20 \\
Prod Chain Ensemble 2       & 69.13  & 72.38 \\
\hline
\end{tabular}
\end{table}

\subsection{Ideology}

As shown in Table \ref{ideo-performance-comparison}, Production Chain 1 achieves an F1 score that surpasses our best individual LLM (Llama) by 2.23 points. Furthermore, shown in Table \ref{Results}, we demonstrate significant effectiveness in generating more reliable and consistent results, evidenced by a standard deviation in F1 performance that is less than half of that observed in the set of LLMs evaluated. Notably, our ensemble approach also outperforms the results published in \cite{ziems24}, which employed simple prompting in a zero-shot format across multiple LLMs. Production Chain 2 shows a lower bound of expected performance when including a model that performs significantly worse than others in the chain.

\begin{table}
\centering
\caption{Ideology Detection Results}
\label{ideo-performance-comparison}
\begin{tabular}{lcc}
\hline
\textbf{Category} & \textbf{Rand. Forward F1} & \textbf{Chain Ensemble F1} \\
\hline
Best Single LLM (GPT)       & -      & 60.85 \\
Worst Single LLM (Phi)      & -      & 44.50 \\
Prod Chain Ensemble 1       & 60.33  & 62.56 \\
Prod Chain Ensemble 2       & 52.76  & 56.50 \\
\hline
\end{tabular}
\end{table}

\subsection{Misinformation}

In line with the findings from the ideology detection task, our experiments in misinformation detection reveal marginal, yet noteworthy benefits when deploying optimally ordered chains compared to the highest-performing individual LLM. The use of Production Chain 1 results in an F1 score improvement of 1.89 points as shown in Table \ref{misinfo-performance-comparison}.

Crucially, our findings indicate a substantial reduction in the standard deviation of performance metrics, pointing to enhanced robustness and consistency in predictions when using our chain ensembling approach. The rank-based chain ensemble reduces the standard deviation in model performance nearly three times, highlighting its capacity to stabilize outputs across diverse data inputs. Production Chain 2 once again highlights the the ability to mitigate the impact of poor performing models within the chain ensemble, outperforming the worst LLM included by nearly 20 F1 points. 

\begin{table}
\centering
\caption{Misinformation Detection Results}
\label{misinfo-performance-comparison}
\begin{tabular}{lcc}
\hline
\textbf{Category} & \textbf{Rand. Forward F1} & \textbf{Chain Ensemble F1} \\
\hline
Best Single LLM (llama)     & -      & 79.01 \\
Worst Single LLM (Phi)      & -      & 54.94 \\
Prod Chain Ensemble 1       & 78.21  & 80.90 \\
Prod Chain Ensemble 2       & 70.68  & 74.57 \\
\hline
\end{tabular}
\end{table}

\section{Discussion}

To the best of our knowledge, our approach has achieved improvements over all available zero-shot benchmarks for stance detection, misinformation identification, and ideology detection. While some of these advancements can be partially attributed to leveraging a more robust LLM, specifically GPT-4o, our chaining method further enhanced the baseline performance of even the best-performing LLM within the chain. Additionally, our rank-based ensemble method not only improved results, but also significantly reduced performance variance. The demonstrated improvement across both Production Chain designs with our chain ensemble methodology over random data forwarding also shows the utility of using confidence thresholds in selecting which data to annotate with which chain link. Moreover, the consistent increase in performance with our data forwarding strategy combined with rank-based ensembling, shown in Figure \ref{chain_size} and Table \ref{Results}, demonstrates a reliable methodology to maximize performance and robustness of zero-shot data annotation systems.

It is important to note that identifying the optimal LLM, prompting scheme, or chain permutation often requires testing a prohibitively large number of models or combinations. In contrast, our method is capable of producing benchmark results without significant fine-tuning or experimentation. Every permutation of models that perform comparably (such as LLaMA, Flan, and GPT) within our rank-based ensemble outperforms the baseline performance of the highest-performing individual LLM. This indicates that data can be processed through any like-performing chain permutation to optimize for speed or cost efficiency while producing performance that exceeds that of a single LLM included in the chain.

The combination of reduced cost and increased performance through a straightforward system implementation represents a crucial finding that could promote broader use of LLMs in data annotation tasks. Furthermore, we believe that the ability to significantly mitigate the impact of poor performing LLMs and boost the performance of highly performant LLMs makes the implementation of a rank-based ensemble chain valuable, particularly in scenarios where model performance is difficult to monitor or performance for a particular task is \textit{a priori} unknown and the optimal models are not clearly defined. This approach provides a robust mechanism to improve the overall reliability of the system by reducing the adverse effects of incorporating sub-optimal models.

Finally, this research also demonstrates the maxim that not all data points are equally hard to label within the same task. This maxim holds not only for benchmark datasets like those tested here, but, we argue, more generally for any given annotation scenario within CSS. Our proposed method exploits this maxim by design, and so should always present a method for economizing resources when annotating data for any given CSS task.

\section{Conclusion}

This study introduces an LLM chain rank-based ensemble approach for data annotation that addresses the critical challenges of scalability, accuracy, and cost-efficiency in labeling large-scale datasets. By aligning multiple LLMs in a predictive chain and routing data based on classification uncertainty, our methodology leverages the strengths of individual models within a cohesive system. This approach not only achieves high performance across various zero-shot text classification tasks in computational social science but also does so at a fraction of the cost of current methods, which rely heavily on prompt engineering and extensive model use, per example to be labeled.

Our methodology also reduces the variability inherent in the outputs of individual LLMs, ensuring that the final classification decisions are more reliable and stable, which is particularly advantageous in practical implementations where zero-shot settings are often employed without the benefit of extensive fine-tuning or task-specific data. The adaptability and interoperability of the chaining system allows it to maintain high performance levels across various conditions as LLMs and prompt engineering continue to evolve.

\subsection{Limitations and Future Works}
Looking forward, future work could explore the dynamic tuning of forwarding thresholds and chain configurations to further enhance performance and adaptability. This would mitigate the limitation of having to batch label data to produce the confidence thresholds our data forwarding strategy depends on, greatly reducing the time to label for a complete dataset. Additionally, investigating the application of this methodology across a broader range of tasks and domains could validate its generalizability and effectiveness beyond the contexts tested in this study. Lastly, we believe further improvements can be implemented in our rank-based ensemble algorithm to increase comparability between models in later chain links. Ultimately, the LLM chain ensemble method offers a promising path forward for scalable, cost-effective, and accurate data annotation, aligning with the growing need for robust AI-driven solutions in the era of big data.

\subsection{Availability}

The authors of this work believe in open-source and reproducible methodology. The open-source models used are available on huggingface at \url{huggingface.co/meta-llama/Meta-Llama-3.1-8B-Instruct}, \url{huggingface.co/mistralai/Mistral-7B-Instruct-v0.3}, \url{huggingface.co/google/flan-ul2}, and \url{huggingface.co/microsoft/Phi-3-medium-128k-instruct}. We access GPT-4o through the OpenAI API \url{https://platform.openai.com/}. All code, prompts, and datasets used in this research project are available \url{github.com/davidthfarr/ChainEnsembles}.




\bibliographystyle{IEEEtran}
\bibliography{refs.bib}
%

\end{document}